\documentclass[11pt,a4paper]{article}
\usepackage[hyperref]{acl2021}
\usepackage{times}
\usepackage{booktabs}
 \usepackage{multirow}
\usepackage{hyperref}
\usepackage{latexsym}
\usepackage{mathrsfs}

\usepackage{microtype}
\usepackage[]{todonotes}
\usepackage{amsmath,amssymb,mathtools,bm,etoolbox}

\usepackage{algorithm}
\usepackage[noend]{algpseudocode}

\aclfinalcopy 
\def\aclpaperid{3266} 

\title{Adapting High-resource NMT Models to Translate Low-resource Related Languages without Parallel Data}

\author{Wei-Jen Ko$^1$\thanks{~~This work was conducted while author was working at Facebook AI} , Ahmed El-Kishky$^{2*}$, Adithya Renduchintala$^3$, Vishrav Chaudhary$^3$, \\
\bf{Naman Goyal$^3$,
Francisco Guzm\'an$^3$, Pascale Fung$^4$, Philipp Koehn$^5$, Mona Diab$^3$} \\
  $^1$University of Texas at Austin, $^2$Twitter Cortex, $^3$Facebook AI\\
  $^4$The Hong Kong University of Science and Technology, $^5$Johns Hopkins University \\
  \texttt{wjko@utexas.edu, aelkishky@twitter.com}\\
  \texttt{\{adirendu,vishrav,naman,fguzman,mdiab\}@fb.com}\\
  \texttt{pascale@ece.ust.hk, phi@jhu.edu}
  }

\date{}

\begin{document}

\maketitle
\begin{abstract}
The scarcity of parallel data is a major obstacle for training high-quality machine translation systems for low-resource languages. Fortunately, some low-resource languages are linguistically related or similar to high-resource languages; these related languages may share many lexical or syntactic structures. In this work, we exploit this linguistic overlap to facilitate translating to and from a low-resource language with only monolingual data, in addition to any parallel data in the related high-resource language. Our method, NMT-Adapt, combines denoising autoencoding, back-translation and adversarial objectives to utilize monolingual data for low-resource adaptation. We experiment on 7 languages from three different language families and show that our technique significantly improves translation into low-resource language compared to other translation baselines.
\end{abstract}

\section{Introduction}
While machine translation (MT) has made incredible strides due to the advent of deep neural machine translation (NMT)~\cite{sutskever2014sequence,bahdanau2014neural} models, this improvement has been shown to be primarily in well-resourced languages with large available parallel training data.

However with the growth of internet communication and the rise of social media, individuals worldwide have begun communicating and producing content in their native low-resource languages. Many of these low-resource languages are closely related to a high-resource language. One such example are ``dialects": variants of a language traditionally considered oral rather than written. Machine translating dialects using models trained on the formal variant of a language (typically the high-resource variant which is sometimes considered the ``standardized form'') can pose a challenge due to the prevalence of non standardized spelling as well significant slang vocabulary in the dialectal variant. Similar issues arise from translating a low-resource language using a related high-resource model (e.g., translating Catalan with a Spanish MT model).

While an intuitive approach to better translating low-resource related languages could be to obtain high-quality parallel data. This approach is often infeasible due to lack specialized expertise or bilingual translators. The problems are exacerbated by issues that arise in quality control for low-resource languages~\cite{Flores}. This scarcity motivates our task of learning machine translation models for low-resource languages while leveraging readily available data such as parallel data from a closely related language or monolingual data in the low-resource language.\footnote{We use low-resource language and dialect or variant interchangeably.}

The use of monolingual data when little to no parallel data is available has been investigated for machine translation. A few approaches involve synthesising more parallel data from monolingual data using backtranslation~\cite{sennrich2015improving} or mining parallel data from large multilingual corpora~\cite{tran2020cross, el2020searching,el2020massive,schwenk2019ccmatrix}. We introduce NMT-Adapt, a zero resource technique that does not need parallel data of any kind on the low resource language.

We investigate the performance of NMT-Adapt at translating two directions for each low-resource language: (1) low-resource to English and (2) English to low-resource. We claim that translating into English can be formulated as a typical unsupervised domain adaptation task, with the high-resource language as the source domain and the related low-resource, the target domain. We then show that adversarial domain adaptation can be applied to this related language translation task. For the second scenario, translating into the low-resource language, the task is more challenging as it involves unsupervised adaptation of the generated output to a new domain. To approach this task, NMT-Adapt jointly optimizes four tasks to perform low-resource translation: (1) denoising autoencoder (2) adversarial training (3) high-resource translation and (4) low-resource backtranslation. 

We test our proposed method and demonstrate its effectiveness in improving low-resource translation from three distinct families: (1) Iberian languages, (2) Indic languages, and (3)  Semitic languages, specifically Arabic dialects. We make our code and resources publicly available.\footnote{\url{https://github.com/wjko2/NMT-Adapt}}

\section{Related Work}

\paragraph{Zero-shot translation}
Our work is closely related to that of zero-shot translation~\cite{googlezs,chenzs,shedivatzs}. However, while zero-shot translation translates between a language pair with no parallel data, there is an assumption that both languages in the target pair have some parallel data with other languages. As such, the system can learn to process both languages. In one work,~\newcite{Currey} improved zero-shot translation using monolingual data on the pivot language. However, in our scenario, there is no parallel data between the low-resource language and any other language. In other work, \newcite{Arivazhaganzs} showed that adding adversarial training to the encoder output could help zero shot training. We adopt a similar philosophy in our multi-task training to ensure our low-resource target is in the same latent space as the higher-resource language.

\paragraph{Unsupervised translation}
A related set of work is the family of unsupervised translation techniques; these approaches translate between language pairs with no parallel corpus of any kind. In work by \newcite{artetxeumt,lampleumt}, unsupervised translation is performed by training denoising autoencoding and backtranslation tasks concurrently. In these approaches, multiple pretraining methods were proposed to better initialize the model~\cite{lampleumt2,XLM,mBART,MASS}.

Different approaches were proposed that used parallel data between X-Y to improve unsupervised translation between X-Z~\cite{GarciaXBT,LiXBT,WangXBT}. This scenario differs from our setting as it does not assume that Y and Z are similar languages. These approaches leverage a cross-translation method on a multilingual NMT model where for a parallel data pair ($S_x$,$S_y$), they translate $S_x$ into language Z with the current model to get $S'_z$. Then use ($S_y$,$S'_z$) as an additional synthesized data pair to further improve the model. \newcite{GarciaXBT2} experiment using multilingual cross-translation on low-resource languages with some success. While these approaches view the parallel data as auxiliary, to supplement unsupervised NMT, our work looks at the problem from a domain adaptation perspective. We attempt to use monolingual data in Z to make the supervised model trained on X-Y generalize to Z.

\paragraph{Leveraging High-resource Languages to Improve Low-resource Translation}
Several works have leveraged data in high-resource languages to improve the translation of similar low-resource languages. \newcite{NeubigLRL} showed that it is beneficial to mix the limited parallel data pairs of low-resource languages with high-resource language data. \newcite{LakewLRL} proposed selecting high-resource language data with lower perplexity in the low-resource language model. \newcite{XiaLRL} created synthetic sentence pairs by unsupervised machine translation, using the high-resource language as a pivot. However these previous approaches emphasize translating from the low-resource language to English, while the opposite direction is either unconsidered or shows poor translation performance. \newcite{Siddhant} trained multilingual translation and denoising simultaneously, and showed that the model could translate languages without parallel data into English near the performance of supervised multilingual NMT. 

\paragraph{Similar language translation} Similar to our work, there have been methods proposed that leverage similar languages to improve translation. \newcite{hassan} generated synthetic English-dialect parallel data from English-main language corpus. However, this method assumes that the vocabulary in the main language could be mapped word by word into the dialect vocabulary, and they calculate the corresponding word for substitution using localized projection. This approach differs from our work in that it relies on the existence of a seed bilingual lexicon to the dialect/similar language. Additionally, the approach only considers translating from a dialect to English and not the reverse direction. Other work trains a massively multilingual many-to-many model and demonstrates that high-resource training data improves related low-resource language translation~\cite{fan2020beyond}. In other work, \newcite{lakewsim} compared ways to model translations of different language varieties, in the setting that parallel data for both varieties is available, the variety for some pairs may not be labeled. Another line of work focus on translating between similar languages. In one such work, \newcite{Pourdamghani} learned a character-based cipher model. In other work, \newcite{wansim} improved unsupervised translation between the main language and the dialect by separating the token embeddings into pivot and private parts while performing layer coordination.
\section{Method}

We describe the NMT-Adapt approach to translating a low-resource language into and out of English without utilizing any low-resource language parallel data. In Section~\ref{sec:eng2lr}, we describe how NMT-Adapt leverages a novel multi-task domain adaptation approach to translating English into a low-resource language. In Section~\ref{sec:lr2eng}, we then describe how we perform source-domain adaptation to translate a low-resource language into English. Finally, in Section~\ref{sec:iterative}, we demonstrate how we can leverage these two domain adaptations, to perform iterative backtranslation -- further improving translation quality in both directions. 

\begin{figure*}
\begin{center}
 \includegraphics[width=0.9\textwidth]{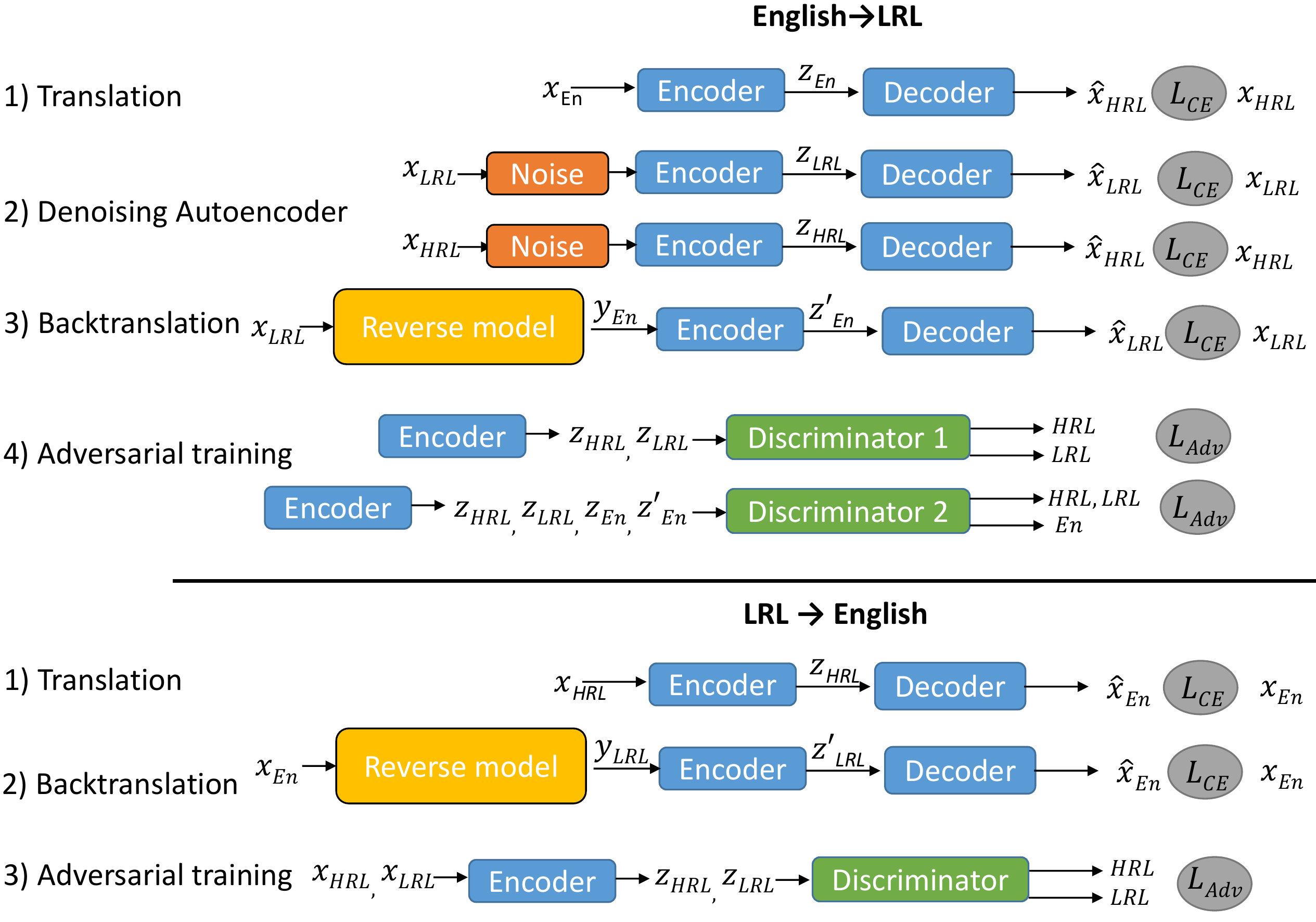}

\end{center}

\caption{Illustration of the training tasks for translating from English into a low-resource language (LRL) and from an LRL to English. }
\vspace{-1.0em}
\label{fig:collection}
\end{figure*}

\subsection{English to Low-resource}
\label{sec:eng2lr}
To translate from English into a low-resource language, NMT-Adapt is initialized with a pretrained mBART model whose pretraining is described in~\cite{mBART}. Then, as shown in Figure~\ref{fig:collection}, we continue to train the model simultaneously with \textit{four} tasks inspired by \cite{lampleumt} and update the model with a weighted sum of the gradients from different tasks. 

The language identifying tokens are placed at the same position as in mBART. For the encoder, \textit{both} high and low-resource language source text, with and without noise, use the language token of the high-resource language [HRL] in the pretrained mBART. For the decoder, the related high and low-resource languages use their own, \textit{different}, language tokens. We initialize the language token embedding of the low-resource language with the embedding from the high-resource language token.
\ \\
\noindent
\textbf{Task 1: Translation} The first task is translation from English into the high-resource language (HRL) which is trained using readily available high-resource parallel data. This task aims to transfer high-resource translation knowledge to aid in translating into the low-resource language. We use the cross entropy loss formulated as follows:
\begin{equation}
\mathcal{L}_{t}=\mathcal{L}_{CE}(\mathcal{D}(Z_{En},[HRL]),X_{HRL})
\end{equation}, where $Z_{En}=\mathcal{E}({X}_{En},[En])$. (${X}_{En},X_{HRL})$ is a parallel sentence pair. $\mathcal{E},\mathcal{D}$ denotes the encoder and decoder functions, which take (input, language token) as parameters. $\mathcal{L}_{CE}$ denotes the cross entropy loss.

\ \\
\noindent
\textbf{Task 2: Denoising Autoencoding} 
For this task, we leverage monolingual text by introducing noise to each sentence, feeding the noised sentence into the encoder, and training the model to generate the original sentence. The noise we use is similar to ~\cite{lampleumt}, which includes a random shuffling and masking of words. The shuffling is a random permutation of words, where the position of words is constrained to shift at most 3 words from the original position. Each word is masked with a uniform probability of 0.1.  This task aims to learn a feature space for the languages, so that the encoder and decoder could transform between the features and the sentences. This is especially necessary for the low-resource language if it is not already pretrained in mBART. Adding noise was shown to be crucial to translation performance in \cite{lampleumt}, as it forces the learned feature space to be more robust and contain high-level semantic knowledge. 

We train the denoising autoencoding on both the low-resource and related high-resource languages and compute the loss as follows:

\begin{equation}
\mathcal{L}_{da}=\sum_{i=LRL,HRL}\mathcal{L}_{CE}(\mathcal{D}(Z_{i},[i]),X_{i})
\end{equation}
, where $Z_{i}=\mathcal{E}(\mathcal{N}({X}_{i}),[HRL])$. ${X}_{i}$ is from the monolingual corpus. 

\ \\
\noindent
\textbf{Task 3: Backtranslation} For this task, we train on English to low-resource backtranslation data. The aim of this task is to capture a language-modeling effect in the low-resource language. We describe how we obtain this data using the high-resource translation model to bootstrap backtranslation in Section~\ref{sec:iterative}.

The objective used is,
\begin{equation}
\mathcal{L}_{bt}=\mathcal{L}_{CE}(\mathcal{D}(Z'_{En},[LRL]),X_{LRL})
\end{equation}
, where $Z'_{En}=\mathcal{E}({Y}_{En},[En])$. (${Y}_{En},X_{LRL})$ is an English to low-resource backtranslation pair.

\ \\
\noindent
\textbf{Task 4: Adversarial Training} The final task aims to make the encoder output language-agnostic features. The representation is language agnostic to the noised high and low-resource languages as well as English. Ideally, the encoder output should contain the semantic information of the sentence and little to no language-specific information. This way, any knowledge learned from the English to high-resource parallel data can be directly applied to generating the low-resource language by simply switching the language token during inference, without capturing spurious correlations~\cite{guzs}.

To adversarially mix the latent space of the encoder among the three languages, we use two critics (discriminators). The critics are recurrent networks to ensure that they can handle variable-length text input. Similar to~\newcite{dwae}, the adversarial component is trained using a Wasserstein loss, which is the difference of expectations between the two types of data. This loss minimizes the earth mover's distance between the distributions of different languages. We compute the loss function as follows:
\begin{equation}
\mathcal{L}_{adv1}=  \mathbb{E}[Disc(Z_{HRL})]-\mathbb{E}[Disc(Z_{LRL})]
\label{eq:adv1}
\end{equation}
\vspace{-1.5em}
\begin{equation}
\mathcal{L}_{adv2}=  \mathbb{E}[Disc(Z_{HRL}\cup Z_{LRL})]
\nonumber
\end{equation}
\vspace{-1.5em}
\begin{equation}
-\mathbb{E}[Disc(Z_{En}\cup Z'_{En})]
\label{eq:adv2}
\end{equation}
 As shown in Equation~\ref{eq:adv1}, the first critic is trained to distinguish between the high and low-resource languages. Similarly, in Equation~\ref{eq:adv2}, the second critic is trained to distinguish between English and non-English (both high, and low-resource languages).

\ \\
\noindent
\textbf{Fine-tuning with Backtranslation}: Finally, we found that after training with the four tasks concurrently, it is beneficial to fine-tune solely using backtranslation for one pass before inference. We posit that this is because while spurious correlations are reduced by the adversarial training, they are not completely eliminated and using solely the language tokens to control the output language is not sufficient. By fine-tuning on backtranslation, we are further adapting to the target side and encouraging the output probability distribution of the decoder to better match the desired output language. 

\subsection{Low-resource to English}
\label{sec:lr2eng}
We propose to model translating from the low-resource language to English as a domain adaptation task and design our model based on insights from domain-adversarial neural network (DANN) \cite{ganin}, a domain adaptation technique widely used in many NLP tasks. This time, we train three tasks simultaneously:

\ \\
\noindent
\textbf{Task 1: Translation} We train high-resource to English translation on parallel data with the goal of adapting this knowledge to translate low-resource sentences. We compute this loss as follows:
\begin{equation}
\mathcal{L}_{t}=\mathcal{L}_{CE}(\mathcal{D}(Z_{HRL},[En]),X_{En})
\end{equation}, where $Z_{HRL}=\mathcal{E}({X}_{HRL},[HRL])$.

\ \\
\noindent
\textbf{Task 2: Backtranslation} Low-resource to English backtranslation translation, which we describe in Section~\ref{sec:iterative}. The objective is as follows:
\begin{equation}
\mathcal{L}_{t}=\mathcal{L}_{CE}(\mathcal{D}(Z'_{LRL},[En]),X_{En})
\end{equation}, where $Z'_{LRL}=\mathcal{E}({Y}_{LRL},[HRL])$.

\ \\
\noindent
\textbf{Task 3: Adversarial Training} We feed the sentences from the monolingual corpora of the high- and low-resource corpora into the encoder, and the encoder output is trained so that its input language cannot be distinguished by a critic. The goal is to encode the low-resource data into a shared space with the high-resource, so that the decoder trained on the translation task can be directly used. No noise was added to the input, since we did not observe an improvement. There is only one recurrent critic, which uses the Wasserstein loss and is computed as follows:
\begin{equation}
\mathcal{L}_{adv}=  \mathbb{E}[Disc(Z_{HRL})]-\mathbb{E}[Disc(Z_{LRL})]
\end{equation}, where $Z_{LRL}=\mathcal{E}({X}_{LRL},[HRL])$.

Similar to the reverse direction, we initialize NMT-Adapt with a pretrained mBART, and use the same language token for high-resource and low-resource in the encoder.

\subsection{Iterative Training}
\label{sec:iterative}
We describe how we can alternate training into/out-of English models to create better backtranslation data improving overall quality.
\begin{algorithm}
\small
\caption{Iterative training}\label{alg:iterative}
\begin{algorithmic}[1]
\State $M^{LRL \text{→}En}_{0} \gets \text{Train HRL to En model}$
\State $X_{mono} \gets \text{Monolingual LRL corpus}$
\State $X_{En} \gets \text{English sentences in the En-HRL parallel corpus}$
\For {k in 1,2...}
\State // Generate backtranslation pairs
\State $\text{Compute} M^{LRL \text{→}En}_{k-1}(X_{mono})$
\State 
\State // Train model as in Sec $3.1$
\State $M^{En\text{→}LRL}_{k} \gets \text{trained En to LRL model}$
\State 
\State // Generate backtranslation pairs
\State $\text{Compute} M^{En \text{→}LRL}_{k}(X_{En})$
\State
\State // Train model as in Sec $3.2$
\State $M^{LRL\text{→}En}_{k} \gets \text{trained LRL to En model}$
\State 
\If{Converged} {break;}
\EndIf
\EndFor

\end{algorithmic}
\end{algorithm}

The iterative training process is described in Algorithm~\ref{alg:iterative}.  We first create English to low-resource backtranslation data by fine-tuning mBART on the high-resource to English parallel data. Using this model, we translate monolingual low-resource text into English treating the low-resource sentences as if they were in the high-resource language. The resulting sentence pairs are used as backtranslation data to train the first iteration of our English to low-resource model.

After training English to low-resource, we use the model to translate the English sentences in the English-HRL parallel data into the low-resource language, and use those sentence pairs as backtranslation data to train the first iteration of our low-resource to English model.

We then use the first low-resource to English model to generate backtranslation pairs for the second English to low-resource model. We iteratively repeat this process of using our model of one direction to improve the other direction.

\section{Experiments}
\subsection{Datasets}
\begin{table*}
\centering
\scriptsize
\setlength{\tabcolsep}{0.5em}
\begin{tabular}{lllrlrrr}
  \toprule
  \bf{Language}&\bf{Group}&\bf{Training Set}&\bf{Train-Size} &\bf{Test Set}&\bf{Test-size}&\bf{Monolingual}&\bf{Mono-Size}\\ 
  \midrule
  Spanish&Iberian&QED~\cite{QED} & 694k & N/A & - &CC-100 & 1M\\
  Catalan&Iberian&N/A& - &Global Voices~\cite{GV} & 15k &CC-100 &1M\\
  Portuguese&Iberian&N/A& - & TED~\cite{TED} & 8k&CC-100 & 1M\\
  \midrule
  Hindi&Indic& IIT Bombay~\cite{IITB} & 769k &N/A & - &CC-100 & 1M\\
  Marathi&Indic& N/A& - &TICO-19~\cite{TICO} & 2k &CC-100 & 1M\\
  Nepali&Indic& N/A& - & FLoRes~\cite{Flores} &3k& CC-100 & 1M\\
  Urdu&Indic& N/A& - &TICO-19~\cite{TICO} & 2k &CC-100 & 1M\\
  \midrule
  MSA&Arabic&QED~\cite{QED} & 465k& N/A & - &CC-100 & 1M\\
  Egyptian Ar.&Arabic&  N/A & - &Forum~\cite{Egy} & 11k &CC-100 & 1.2M\\
  Levantine Ar.&Arabic&  N/A& - &Web text~\cite{Lev} & 11k &CC-100 & 1M\\
\bottomrule
  
\end{tabular}
\caption{The sources and size of the datasets we use for each language. The HRLs are used for training and the LRLs are used for testing.}
\vspace{-1.0em}
\label{tab:5}
\end{table*}

We experiment on three groups of languages. In each group, we have a large quantity of parallel training data for one language(high-resource) and no parallel for the related languages to simulate a low-resource scenario. 

Our three groupings include (i) \textit{Iberian languages}, where we treat Spanish as the high-resource and Portuguese and Catalan as related lower-resource languages. (ii) \textit{Indic languages} where we treat Hindi as the high-resource language, and Marathi, Nepali, and Urdu as lower-resource related languages (iii) \textit{Arabic}, where we treat Modern Standard Arabic (MSA) as the high-resource, and Egyptian and Levantine Arabic dialects as low-resource. Among the languages, the relationship between Urdu and Hindi is a special setting; while the two languages are mutually intelligible as spoken languages, they are written using different scripts. Additionally, in our experimental setting, all low-resource languages except for Nepali were not included in the original mBART pretraining.

The parallel corpus for each language is described in Table \ref{tab:5}. Due to the scarcity of any parallel data for a few low-resource languages, we are not able to match the training and testing domains. For monolingual data, we randomly sample 1M sentences for each language from the CC-100 corpus\footnote{\url{http://data.statmt.org/cc-100/}}~\cite{XLMR,wenzek-etal-2020-ccnet}. For quality control, we filter out sentences if more than $40\%$ of characters in the sentence do not belong to the alphabet set of the language. For quality and memory constraints, we only use sentences with length between $30$ and $200$ characters.

\ \\
\noindent
\textbf{Collecting Dialectical Arabic Data} While obtaining low-resource monolingual data is relatively straightforward, as language identifiers are often readily available for even low-resource text~\cite{jauhiainen2019automatic}, identifying dialectical data is often less straightforward. This is because many dialects have been traditionally considered oral rather than written, and often lack standardized spelling, significant slang, or even lack of mutual intelligibility from the main language. In general, dialectical data has often been grouped in with the main language in language classifiers.

We describe the steps we took to obtain reliable dialectical Arabic monolingual data. As the CC-100 corpus does not distinguish between Modern Standard Arabic (MSA) and its dialectical variants, we train a finer-grained classifier that distinguishes between MSA and specific colloquial dialects. We base our language classifier on a BERT model pretrained for Arabic~\cite{arabicBERT} and fine-tune it for six-way classification: (i) Egyptian, (ii) Levantine, (iii) Gulf, (iv) Maghrebi,  (v) Iraqi dialects as well as (vi) the literary Modern Standard Arabic (MSA).  We use the data from \cite{MADAR} and \cite{AOC} as training data, and the resulting classifier has an accuracy of $91$\% on a held-out set. We take our trained Arabic dialect classifier and further classify Arabic monolingual data from CC-100 and select MSA, Levantine and Egyptian sentences as Arabic monolingual data for our experiments.

\subsection{Training Details}
We use the RMSprop optimizer with learning rate $0.01$ for the critics and the Adam optimizer for the rest of the model. We train our model using eight GPUs and a batch size of $1024$ tokens per GPU. We update the parameters once per eight batches. For the adversarial task, the generator is trained once per three updates, and the critic is trained every update. 

Each of the tasks of (i) translation, (ii) backtranslation as well as (iii) LRL and HRL denoising (only for En$\rightarrow$LRL direction), have the same number of samples and their cross entropy loss has equal weight. The adversarial loss, $\mathcal{L}_{adv}$, has the same weight on the critic, while it has a multiplier of $-60$ on the generator (encoder). This multiplier was tuned to ensure convergence and is negative as it's opposite to the discriminator loss.

\begin{table*}
\centering
\scriptsize
\setlength{\tabcolsep}{0.5em}
\begin{tabular}{llrrrrrr}
    \toprule
    \emph{En$\rightarrow$ LRL} &&\bf{Un-adapted Model}& \multicolumn{4}{c}{ \bf{Adapted Models}} \\\cmidrule(r){3-3}\cmidrule(l){4-7}
  \bf{LRL}& \bf{HRL}& En$\rightarrow$HRL  & Adv & BT& BT+Adv& BT+Adv+fine-tune\\
  \midrule
  Portuguese&Spanish & 3.8 &10.1& 14.8&18.0&\textbf{21.2} \\
  Catalan&Spanish & 6.8 &9.1& 21.2&22.5&\textbf{23.6} \\
  Marathi&Hindi & 7.3 &8.4& 9.5&15.6&\textbf{16.1} \\
  Nepali&Hindi & 11.2 &17.6& 16.7&25.3&\textbf{26.3} \\
  Urdu&Hindi & 0.3 &3.4&0.2& \textbf{7.2} & -\\
  Egyptian Arabic&MSA& 3.5 & 3.8 &\textbf{8.0}& \textbf{8.0}&\textbf{8.0} \\
 Levantine Arabic&MSA& 2.1 & 2.1 &4.8& \textbf{5.1}&4.7 \\
\bottomrule
\end{tabular}
\caption{BLEU score of the first iteration on the English to low-resource direction. Both the adversarial (Adv) and backtranslation (BT) components contribute to improving the results. The fine-tuning step is omitted for Urdu as decoding is already restricted to a different script-set from the related high-resource language.}
\label{tab:1}
\end{table*}

\begin{table*}
\centering
\scriptsize

\setlength{\tabcolsep}{0.5em}
\begin{tabular}{llrrrrrr}
    \toprule
\emph{LRL$\rightarrow$En}    &&\bf{Un-adapted Model}& \multicolumn{3}{c}{ \bf{Adapted Models}} \\\cmidrule(r){3-3}\cmidrule(l){4-6}
  \bf{LRL}& \bf{HRL}& HRL$\rightarrow$En  & Adv & BT& BT+Adv\\
  \midrule

  Portuguese&Spanish & 12.3 &21.7& 32.7&\textbf{36.0} \\
  Catalan&Spanish &12.2& 13.9&\textbf{25.3}&24.6 \\
  Marathi&Hindi & 3.9& 7.0&8.1&\textbf{12.7} \\
  Nepali&Hindi & 14.8 &16.9& 14.1&\textbf{18.2} \\
  Urdu&Hindi &0.3  &1.0&\textbf{10.5} &\textbf{10.5} \\
  Egyptian Arabic&MSA& 14.9 & 14.0 &15.2& \textbf{15.8} \\
 Levantine Arabic&MSA& \textbf{9.3} & 6.7 &\textbf{9.3}& 9.0 \\
\bottomrule
\end{tabular}
\caption{BLEU score of the first iteration on the LRL to English direction. Both the adversarial(Adv) and backtranslation (BT) components contribute to improving the results.}
\vspace{-1.0em}
\label{tab:2}
\end{table*}

For the first iteration, we train $128$ epochs from English to the low-resource language and $64$ iterations from low-resource language to English. For the second iteration we train $55$ epochs for both directions. We follow the setting of~\cite{mBART} for all other settings and training parameters.

The critics consist of four layers: the third layer is a  bidirectional GRU and the remaining three are fully connected layers. The hidden layer sizes are $512$, $512$ and $128$ and we use an SELU activation function.

We ran experiments on 8-GPUs. Each iteration took less than 3 days and we used publicly available mBART-checkpoints for initialization.  GPU memory usage of our method is only slightly larger than mBART. While we introduce additional parameters in discriminators, these additional parameters are insignificant compared to the size of the mBART model.

\subsection{Results}
We present results of applying NMT-Adapt to low-resource language translation.

\subsubsection{English to Low-Resource}
We first evaluate performance of translating into the low-resource language. We compare the first iteration of NMT-Adapt to the following baseline systems: \textbf{(i)}  En$\rightarrow$HRL Model: directly using the model trained for En$\rightarrow$HRL translation. \textbf{(ii)} Adversarial: Our full model without using the backtranslation objective and without the final fine-tuning. \textbf{(iii) }Backtranslation: mBART fine-tuned on backtranslation data created using the HRL$\rightarrow$En model. \textbf{(iv)} BT+Adv: Our full model without the final fine-tuning. \textbf{(v)} BT+Adv+fine-tune: Our full model (NMT-Adapt) as described in Section 3.

As seen in Table~\ref{tab:1}, using solely the adversarial component only, we generally see improvement in the BLEU scores over using the high-resource translate model. This suggests that our proposed method of combining denoising autoencoding with adversarial loss is effective in adapting to a new target output domain.

Additionally, we observe a large improvement using only backtranslation data. This demonstrates that using the high-resource translation model to create LRL-En backtranslation data is highly effective for adapting to the low-resource target. 

We further see that combining adversarial and backtranslation tasks further improve over each individually, showing that the two components are complementary. We also experimented on En-HRL translation with backtranslation but without adversarial loss. However, this yielded much worse results, showing that the improvement is not simply due to multitask learning.

For Arabic, backtranslation provides most of the gain, while for Portuguese and Nepali, the adversarial component is more important. For some languages like Marathi, the two components provides small gains individually, but shows a large improvement while combined.

\begin{table*}
\centering
\scriptsize
\setlength{\tabcolsep}{0.5em}
\begin{tabular}{lllllll}
\toprule
   & \multicolumn{3}{c}{\bf{English→LRL}}&\multicolumn{3}{c}{ \bf{LRL→English}}\\\cmidrule(r){2-4}\cmidrule(l){5-7}
  \bf{Language} & NMT-Adapt It.1 & NMT-Adapt It.2 & MBART+MASS& NMT-Adapt It.1 & NMT-Adapt It.2 & MBART+MASS\\
  \midrule
  Portuguese&21.2 &\textbf{30.7}&26.6&36.0&\textbf{39.8}&38.1 \\
  Catalan& 23.6&\textbf{27.2}&23.3&24.6&\textbf{27.7}&22.9 \\
  Marathi& 16.1&\textbf{19.2}&13.1&12.7&\textbf{15.0}&5.8 \\
  Nepali& \textbf{26.3}&\textbf{26.3} &11.9&18.2&\textbf{18.8}&2.1 \\
  Urdu& 7.2&\textbf{14.6}&5.1&10.5&\textbf{13.6}&4.9 \\
  Egyptian Ar.&\textbf{8.0} &6.6&3.3&\textbf{15.8}&-&11.7 \\
  Levantine Ar.&\textbf{5.1} &4.5&1.9&\textbf{9.0}&-&6.0 \\
  
\bottomrule
\end{tabular}
\caption{BLEU results of iterative training. The second iteration generally improves among the first iteration, and NMT-Adapt outperforms the MBART+MASS baseline. For Arabic, as iteration 2 into Arabic was worse than iteration 1, we omit the corresponding iteration 2 into English.}
\label{tab:3}
\vspace{-1.5em}

\end{table*}

\begin{table}
\centering
\scriptsize
\setlength{\tabcolsep}{0.5em}
\begin{tabular}{llll}
\toprule
 & & \multicolumn{2}{c}{\bf{BLEU}}\\\cmidrule{3-4}
  & & En→Ne&Ne→En\\
 \midrule
  \multirow{3}{*}{\shortstack{Unsupervised+\\ Hi parallel}}&NMT-Adapt&\textbf{9.2} &\textbf{18.8} \\
  &\cite{Flores}&8.3 &\textbf{18.8} \\
  &\cite{mBART}&-&17.9 \\
\midrule
 \shortstack{Unsupervised+\\ Multi. parallel} &\cite{GarciaXBT2}& 8.9&21.7 \\
\midrule
\midrule 
   \multirow{2}{*}{\shortstack{Sup. with Hi}}&\cite{Flores}& 8.8&\textbf{21.5} \\
 &\cite{mBART}&\textbf{9.6} &21.3 \\
 \midrule
  Sup. w/o Hi&\cite{Flores}& 4.3&7.6 \\
 \bottomrule
\end{tabular}
\caption{Comparison with previous work on FLoRes dataset. NMT-Adapt outperforms previous unsupervised methods on En→Ne, and achieves similar performance to unsupervised baselines on Ne→En.}
\label{tab:4}
\end{table}

\begin{table}
\centering
\small
\setlength{\tabcolsep}{0.5em}
\begin{tabular}{lr}
\toprule 

\bf{\# sentences}  & \bf{BLEU}\\
\midrule
10k & 11.3\\
100k & 14.1\\
1M & 16.1 \\
\bottomrule
\end{tabular}
\caption{First iteration English to Marathi results with variable amount of monolingual data.}
\label{tab:6}
\vspace{-1.0em}

\end{table}

For Urdu, we found that backtranslation only using the Hindi model completely fails; this is intuitive as Hindi and Urdu are in completely different scripts and using a Hindi model to translate Urdu results in effectively random backtranslation data. When we attempt to apply models trained with the adversarial task, the model generates sentences with mixed Hindi, Urdu, and English. To ensure our model solely outputs Urdu, we restricted the output tokens by banning all tokens containing English or Devanagari (Hindi) characters. This allowed our model to output valid and semantically meaningful translations. This is an interesting result as it shows that our adversarial mixing allows translating similar languages even if they're written in different scripts. We report the BLEU score with the restriction. Since the tokens are already restricted, we skip the final fine-tuning step.

\subsubsection{Low-resource to English}
Table \ref{tab:2} shows the results of the first iteration from translating from a low-resource language into English. We compare the following systems \textbf{(i)} HRL→En model: directly using the model trained for HRL→En translation. \textbf{(ii)} Adversarial: similar to our full model, but without using the backtranslation objective. \textbf{(iii)} Backtranslation: mBART fine-tuned on backtranslation data from our full model in the English-LRL direction. \textbf{(iv)} BT+Adv: Our full model.

For this direction, we can see that both the backtranslation and the adversarial domain adaptation components are generally effective. The exception is Arabic which may be due to noisiness of our dialect classification compared to low-resource language classification. Another reason could be due to the lack of written standardization for spoken dialects in comparison to low-resource, but standardized languages.

For these experiments, we did not apply any special precautions for Urdu on this direction despite it being in a different script from Hindi.

\subsubsection{Iterative Training}
Table \ref{tab:3} shows the results of two iterations of training. For languages other than Arabic dialects, the second iteration generally shows improvement over the first iteration, showing that we can leverage an improved model in one direction to further improve the reverse direction. We found that the improvement after the third iteration is marginal. 

We compare our results with a baseline using the HRL language as a pivot. The baseline uses a fine tuned mBART \cite{mBART} to perform supervised translation between English and the HRL, and uses MASS \cite{MASS} to perform unsupervised translation between the HRL and the LRL. The mBART is tuned on the same parallel data used in our method, and the MASS uses the same monolingual data as in our method. For all languages and directions, our method significantly outperforms the pivot baseline.

\subsubsection{Comparison with Other Methods}
In table 5, we compare a cross translation method using parallel corpora with multiple languages as auxiliary data~\cite{GarciaXBT2} as well as results reported in~\cite{Flores} and \cite{mBART}. All methods use the same test set, English-Hindi parallel corpus, and tokenization for fair comparison.
For English to Nepali, NMT-Adapt outperforms previous unsupervised methods using Hindi or multilingual parallel data, and is competitive with supervised methods. For Nepali to English direction, our method achieves similar performance to previous unsupervised methods.
Note that we use a different tokenization than in table 3 and 4, to be consistent with previous work.
\subsubsection{Monolingual Data Ablation}
Table \ref{tab:6} shows the first iteration English to Marathi results while varying the amount of monolingual data used. We see that the BLEU score increased from $11.3$ to $16.1$ as the number of sentences increased from $10$k to $1$M showing additional monolingual data significantly improves performance.

\section{Conclusion}
We presented NMT-Adapt, a novel approach for neural machine translation of low-resource languages which assumes zero parallel data or bilingual lexicon in the low-resource language. Utilizing parallel data in a similar high resource language as well as monolingual data in the low-resource language, we apply unsupervised adaptation to facilitate translation to and from the low-resource language. Our approach combines several tasks including adversarial training, denoising language modeling, and iterative back translation to facilitate the adaptation. Experiments demonstrate that this combination is more effective than any task on its own and generalizes across many different language groups.

\bibliography{acl2020}
\bibliographystyle{acl_natbib}

\appendix


\end{document}